\documentclass[11pt]{article}

\usepackage[preprint]{acl}

\usepackage{times}
\usepackage{latexsym}

\usepackage[T1]{fontenc}

\usepackage[utf8]{inputenc}

\usepackage{microtype}

\usepackage{inconsolata}

\usepackage{graphicx}

\usepackage{amsmath}
\usepackage{amsfonts}
\usepackage[symbol]{footmisc}

\usepackage{pgfplots}
\pgfplotsset{compat=1.18}
\usepgfplotslibrary{statistics}
\usepgfplotslibrary{colormaps}
\usepackage{tikz}
\usepackage{float}
\usepackage{booktabs}
\usetikzlibrary{arrows, arrows.meta, matrix, positioning, math, external, patterns}

\newsavebox{\predecbox}
\sbox{\predecbox}{\tikz[baseline=-0.55ex]\draw[-{Stealth[length=3.5pt,width=3.5pt]},line width=0.7pt] (0.55em,0) -- (0,0);}
\newsavebox{\indecbox}
\sbox{\indecbox}{\tikz[baseline=-0.55ex]\draw[-{Stealth[length=3.5pt,width=3.5pt]},line width=0.7pt] (0,0.45em) -- (0,-0.35em);}
\newsavebox{\postdecbox}
\sbox{\postdecbox}{\tikz[baseline=-0.55ex]\draw[-{Stealth[length=3.5pt,width=3.5pt]},line width=0.7pt] (0,0) -- (0.55em,0);}
\DeclareRobustCommand{\predec}{\usebox{\predecbox}}
\DeclareRobustCommand{\indec}{\usebox{\indecbox}}
\DeclareRobustCommand{\postdec}{\usebox{\postdecbox}}

\title{Beyond a Global Norm: Personalizing Toxicity Sensitivity\\in Language Models Without Retraining}

\author{
  Rares A.C. Diaconescu*,
  Iulia Slănină*, 
  Alina Florea*,
  Andrei B. Trache*,
  Miruna E. Coroi*,\\
  \textbf{Anne Arzberger, Jie Yang, and Enrico Liscio}
  \\
  Delft University of Technology\\
  \texttt{\{r.a.c.diaconescu, i.slanina-1, a.florea-1, a.b.trache, m.e.coroi,}\\
  \texttt{a.arzberger, j.yang-3, e.liscio\}@tudelft.nl}
}

\begin{document}
\maketitle
\begin{abstract}
    Reducing toxicity is often framed as a global alignment problem, yet perceptions of harmful language are subjective and context-dependent. We present the first comparative evaluation of training-free methods for aligning language generation to user-specific toxicity sensitivities across three inference-time intervention stages: pre-decoding (prompt conditioning and rewriting), in-decoding (token, logit, and representation steering), and post-decoding (candidate re-ranking). Evaluated against toxicity sensitivity targets derived from the PRISM dataset, all methods reduce alignment error by 28–47\%. However, the results reveal a fundamental trade-off between alignment effectiveness, personalization, and general language quality, showing how toxicity sensitivity alignment is an inherently multi-objective problem.
    
\end{abstract}

\footnotetext[1]{Equal contribution.}

\section{Introduction}
LLMs increasingly mediate interactions in which safety is not a property of text alone, but a situated judgement about speaker, audience, and context \citep{arzberger2024nothing}. Toxicity moderation makes this especially visible, as it reaches into forms of speech whose meaning depends on social and pragmatic context \cite{balayn2021automatic}.
Such distinctions are not reliably captured by global toxicity labels, as perceptions of toxicity vary systematically across individuals and communities, reflecting differences in identity, experience, and norms rather than noise \citep{sap2022annotators,kirk2024prism}. Optimising language models toward a single notion of acceptable language risks mismatching users' preferences and communicative contexts, motivating personalised toxicity alignment.

Established alignment approaches, e.g., RLHF \citep{ouyang2022training} and DPO \citep{rafailov2023direct}, typically aggregate heterogeneous judgements into a single training signal. This produces useful general-purpose safety behaviour, but it also converts disagreement into a single optimisation target, typically at the cost of subsuming minority perspectives \citep{plank-2022-problem,sorensen2024roadmap}. 
\textbf{Training-free alignment} offers a promising alternative to accommodate user preferences without training separate personalised models \cite{pan2025survey}. Instead of updating model weights, this approach steers model behaviour at inference time.
Although it has shown promise for tasks such as safety alignment \cite{wang-etal-2024-inferaligner}, style control \cite{scalena-etal-2026-steering}, and improving factuality or helpfulness \cite{lin2024unlocking}, its effectiveness for personalised toxicity alignment has not been systematically evaluated.

This paper presents a comparative evaluation of training-free approaches for personalised toxicity sensitivity alignment. Using the PRISM dataset \citep{kirk2024prism}, we derive per-user toxicity sensitivity profiles and evaluate how different approaches steer model outputs closer to users' profiles without fine-tuning the base model. We compare seven methods, spanning approaches that act before generation by conditioning the prompt (\textit{pre-decoding}), during generation by modifying token probabilities or hidden states (\textit{in-decoding}), or after generation by selecting among generated candidates (\textit{post-decoding}) \cite{pan2025survey}. These stages trade off differently: pre-decoding needs only input access but controls indirectly, in-decoding controls directly but requires model internals, and post-decoding is auditable but bounded by the candidate pool.

Our results show that training-free alignment can effectively steer an off-the-shelf model toward language that reflects an individual's sensitivity to toxicity without retraining: all seven methods reduce the distance to user-specific toxicity targets (by 28\% to 47\%), but score differently across toxicity categories (with the largest gains on \textsc{severe toxicity} and \textsc{threat}). 
Pre-decoding achieves the best alignment, though partly through a broad shift toward safer text and at some cost to knowledge retention, while in- and post-decoding methods concentrate smaller reductions on each user's most sensitive categories. Furthermore, results vary across demographic groups, including one case where the intervention reduces alignment. These findings suggest that toxicity sensitivity alignment is a multi-objective problem, requiring methods and evaluations that jointly account for user preferences, language quality, and subgroup effects.
 
\section{Related Works}
We situate our work across toxicity modelling, mitigation, and training-free model alignment.

\subsection{Toxicity Modelling in NLP}
Toxicity modelling has traditionally focused on identifying harmful language through supervised classification \citep{rottger-etal-2022-multilingual,10.1145/3580494} and reducing toxic outputs from generative models \citep{pozzobon-etal-2023-goodtriever,dan-etal-2026-survey}. Alignment-based methods extend this mitigation agenda by optimising model behaviour against collected human preferences. Techniques such as Reinforcement Learning from Human Feedback (RLHF) \citep{ouyang2022training} and Direct Preference Optimisation (DPO) \citep{rafailov2023direct} have been shown to reduce toxicity and harmfulness \citep{ICLR2024_dd1577af,NEURIPS2024_c182ec59}, improve multilingual detoxification \citep{li-etal-2024-preference}, and increase robustness against toxic interactions across conversations \citep{kim-lee-2024-adversarial}.

However, both toxicity modelling and alignment-based mitigation tend to operationalise harm as a global target: a property to be detected, scored, reduced, or optimised against. This assumption is increasingly challenged by work showing that toxicity judgements are context-dependent and socially situated. Annotations reflect annotators' perspectives and social experiences \citep{sap2022annotators, arzbergerLabel}, vary across instruction paradigms \citep{rottger-etal-2022-two}, and depend on distinctions between different harmful concepts \citep{cercas-curry-etal-2024-subjective}. Hate speech models may therefore inherit the definitions encoded by their training datasets rather than capture an objective notion of harm \citep{khurana-etal-2025-defverify}.

This situatedness poses a practical challenge for dominant alignment pipelines. Because RLHF and DPO typically rely on additional training over aggregated preference data, they produce a single optimised model rather than a lightweight mechanism for adapting toxicity thresholds to individual users or contexts. As a result, they are well suited to general-purpose safety behaviour, but less suited to cases where users differ in what language they consider harmful, acceptable, or necessary in context. This motivates training-free approaches that can steer toxicity behaviour at inference time without retraining model weights.

\subsection{Training-Free Alignment}
\label{sec:related-works:training-free}

Training-free alignment steers model output at inference time without updating weights. We distinguish three intervention stages \citep{pan2025survey}. 

\textbf{Pre-decoding} methods modify the input through prompt conditioning or rewriting before generation \citep{lin2024unlocking, cheng2024black}. They are black-box compatible, but offer indirect personalisation because the model may ignore, reinterpret, or overgeneralise user-specific toxicity thresholds. 

\textbf{In-decoding} methods intervene during generation by re-weighting token probabilities with an external classifier or discriminator \citep{krause2021gedi, yang2021fudge}, or by modifying internal representations through activation steering and representation engineering \citep{wehner2025taxonomy,scalena-etal-2026-steering}, where steering vectors can also be learned directly from preference pairs rather than computed from activation differences \citep{cao2024bipo}. They enable personalisation by shaping generation in real time, but require access to logits, the decoding loop, or model internals, and can degrade fluency when applied too strongly. 

\textbf{Post-decoding} methods leave generation unchanged and filter, re-rank, or select among completed candidates using an external judge, classifier, or user-specific scoring function \citep{cao-etal-2024-defending,NEURIPS2024_a51a74b2,oak-etal-2025-ranking}. They are auditable and black-box compatible, but limit personalisation to the diversity of the generated candidate pool.

These stages offer distinct trade-offs for personalising generation without retraining. Yet it remains unclear how they compare under the same user-specific toxicity sensitivity profiles, and whether intervention improves personalisation or mainly introduces stronger quality and stability costs. In this work, we further describe (Section~\ref{sec:methods}) and evaluate (Sections~\ref{sec:metrics} and \ref{sec:exp-setup}) seven training-free alignment methods: two pre-decoding prompt-level approaches, three in-decoding methods, and a post-decoding re-ranker with two distance matchers.

\section{Comparative Evaluation}
We compare seven training-free approaches to personalised toxicity sensitivity alignment on a shared benchmark. We report personalization alignment as the main evaluation metric, along with knowledge retention and generation quality to measure alignment trade-offs.

\subsection{Personalization Target}
The PRISM dataset \citep{kirk2024prism} links participants, prompts, and ratings, allowing the evaluation of alignment against individual users rather than against a single global toxicity threshold. 
For each user, we construct a toxicity-sensitivity profile using Perspective API\footnote{\url{https://www.perspectiveapi.com/}}, a widely used toxicity classifier that scores text across multiple harm-related attributes. We use the following six toxicity categories: \textsc{toxicity}, \textsc{severe toxicity}, \textsc{identity attack}, \textsc{insult}, \textsc{profanity}, and \textsc{threat}. 

Each dataset record consists of a prompt $p$, a set of available responses $R=\{r_1, \dots, r_n\}$, and the acceptability rating $\rho_{u,r} \in [0,100]$ that user $u$ assigned to response $r$. We denote the user's preferred response to prompt $p$ (i.e., the response with the highest rating) with $r^{\star}_{u,p}$, and the history of all prompts annotated by user $u$ as $P_u$.
First, we convert ratings into dislike weights, as follows:

\[
d_{u,r} = 1 - \frac{\rho_{u,r}}{100},
\]

\noindent so that strongly disliked responses contribute more to the profile. Then, let $\tau_c(r)$ be the Perspective API score of response $r$ on category $c$. The user's sensitivity on category $c$ is computed as the dislike-weighted mean toxicity of the responses they rated:

\begin{equation}
s_{u,c} =
\frac{\sum_{r \in P_u} d_{u,r} \tau_c(r)}
{\sum_{r \in P_u} d_{u,r}}.
\label{eq:profile}
\end{equation}

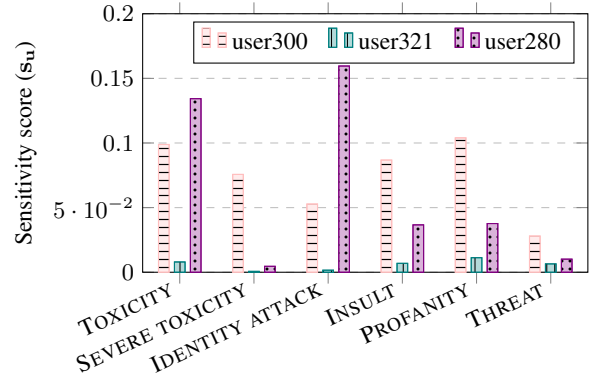
\begin{figure}[t]
\small
\centering
\begin{tikzpicture}
\begin{axis}[
width=0.97\columnwidth,
height=5cm,
ylabel=Sensitivity score ($\mathbf{s_u}$),
ybar,
ymin=0,
ymax=0.2,
xtick={0,1,2,3,4,5},
xticklabels={\textsc{Toxicity}, \textsc{Severe toxicity}, \textsc{Identity attack}, \textsc{Insult}, \textsc{Profanity}, \textsc{Threat}},
x tick label style={rotate=25,anchor=east},
bar width=4pt,
enlarge x limits=0.1,
legend pos=north east,
legend cell align={left},
legend columns=3,
legend style={/tikz/every even column/.append style={column sep=0.2cm}},
ymajorgrids=true,
grid style=dashed,
]
\addplot [pink,fill=pink!30!white,postaction={pattern=horizontal lines}] table[x={point}, y={user300}] {
point user300
0 0.098847
1 0.075681
2 0.052780
3 0.086785
4 0.103958
5 0.028006
};

\addplot [teal,fill=teal!30!white,postaction={pattern=vertical lines}] table[x={point}, y={user321}] {
point user321
0 0.007951
1 0.000681
2 0.001586
3 0.006890
4 0.011256
5 0.006462
};

\addplot [violet,fill=violet!30!white,postaction={pattern=dots}] table[x={point}, y={user280}] {
point user280
0 0.134289
1 0.004590
2 0.159558
3 0.036661
4 0.037677
5 0.010130
};

\legend{user300, user321, user280}
\end{axis}
\end{tikzpicture}
\caption{Toxicity sensitivity scores across the six toxicity categories for three users in the PRISM dataset.}
\label{fig:prism}
\end{figure}

High values indicate toxicity dimensions that the user repeatedly rated down. We use the vector of user sensitivities across the six toxicity categories ($\mathbf{s}_u$) as the personalisation signal used to evaluate our methods. Figure~\ref{fig:prism} shows how such sensitivities widely vary across PRISM users and categories, further motivating the need for personalized alignment. However, we want to stress that, since we infer the profile from automatic toxicity scores rather than directly elicited from users, it should be read as a proxy for user-specific sensitivity, not as a definitive account of perceived harm.

\subsection{Methods}
\label{sec:methods}
We compare seven training-free alignment methods (grouped into the three intervention stages described in Section~\ref{sec:related-works:training-free}) to steer a language model to generate a response aligned with a user's sensitivity. 
We mark each method with the stage at which it intervenes: \predec~for pre-decoding, \indec~for in-decoding, and \postdec~for post-decoding.
Precisely, given a user's sensitivity vector $\mathbf{s_u}$ and a novel prompt $p'$, we steer the model into generating a response $r'_p$ that is aligned with the user's sensitivity (see Section~\ref{sec:metrics} for the evaluation procedure).
In all cases, we avoid using Perspective API as part of the method to ensure the validity of the evaluation procedure.
Appendix~\ref{appendix:implementation-details} provides additional implementation details and the values of the control parameters (e.g., $k$, $\alpha$, and $\gamma$).

\subsubsection{Pre-Decoding (\predec)}

We compare two methods that require only input access to evaluate personalisation effects without access to decoding internals.

\paragraph{\predec~URIAL: Prompt conditioning.}
Untuned LLMs with Restyled In-Context Alignment (URIAL) shows that base models can be substantially aligned through instructions and in-context examples alone, without parameter updates \citep{lin2024unlocking}. We adapt this logic by adding user-specific toxicity sensitivity guidance before the original prompt, preserving the user's request while conditioning the model toward the relevant safety profile. Concretely, the prefix contains a shared safety instruction and examples selected from the user's primary toxicity-sensitivity category, so the target model receives both the original task and a signal about which type of harmful wording should be handled most carefully.

\paragraph{\predec~PBPO: Prompt rewriting.}
Black-box Prompt Optimisation (PBPO) treats the prompt itself as the object of intervention, improving alignment by reformulating the input before it reaches the target model \citep{cheng2024black}. We evaluate a variant that applies this idea to personalised toxicity sensitivity alignment by rewriting the prompt according to the user's six-dimensional toxicity profile. Precisely, a separate model (\texttt{Mistral-7B-Instruct}, in our experiments) receives the original prompt together with the user's six sensitivity labels and returns a neutralized version of the same request, which is then sent to the target model.

\subsubsection{In-Decoding (\indec)}

We compare three methods that steer model behavior by adjusting its hidden states or logits. These methods require access to the model's internals but allow for more transparent control.

\paragraph{\indec~PCAA: activation steering.}
We introduce Personalised Contrastive Activation Addition (PCAA), a confidence-gated hybrid variant of contrastive activation addition \citep{wehner2025taxonomy}. We first compute a population-level toxicity direction in activation space ($v_{\text{pop}}$) from PRISM preference pairs. To personalise this direction, we fit a Bradley-Terry preference model over user choices to obtain a low-rank basis of safety directions together with a per-user coordinate vector $z_u$, from which we compose a user-specific residual $\delta_u$. We combine the two into a confidence-gated hybrid direction:
\[
v_u = v_{\text{pop}} + \gamma\, g_u\, \delta_u,
\qquad
g_u = \min\!\Big(1, \tfrac{\lVert z_u \rVert}{\kappa}\Big),
\]
where $\gamma$ scales the personalised residual and $\kappa$ is a norm threshold above which a user's coordinates are treated as fully reliable.
We then add $\alpha v_u$ to the residual stream to steer hidden states during the forward pass.
The gate $g_u$ scales the personalised residual by how confidently the user's coordinates are estimated, and falls back to the population direction for unknown or low-confidence users ($g_u\!\to\!0$). 

\paragraph{\indec~CGD: Classifier-guided token selection.}
We use Classifier-Guided Decoding (CGD) to perform token-level selection at the decoding step, following prior work such as PPLM~\cite{dathathri2019plug}, GeDi~\cite{krause2021gedi}, and FUDGE~\cite{yang2021fudge}. At each decoding step, we select the top-$k$ next-token candidates ($\xi_1 \dots \xi_k$). We use Detoxify~\cite{detoxify}, a lightweight multi-label toxicity classifier, to score each candidate continuation across the six toxicity categories. The scores $\tau_c(r)$ are weighted by the user's sensitivity vector $\mathbf{s_u}$, so that categories the user is more sensitive to have a stronger effect on token selection:
%
\[
  z'_u(\xi) \;=\; z(\xi) \;-\; \alpha \sum_{c=1}^{6} s_{u,c} \, \tau_c(r),
  \label{eq:guided-logit}
\]
%
where $z(\xi)$ is the original logit and $\alpha$ controls guidance strength. We then choose the token candidate with the highest adjusted logit score.

\paragraph{\indec~EADS: expert--anti-expert steering.}
Expert--Anti-Expert Differential Steering (EADS) 
shifts the next-token distribution by contrasting desirable and undesirable continuations \citep{liu2021dexperts,liu2024proxy}. 
It runs three synchronised decoding branches: a base branch, an expert branch conditioned to model the desired response style, and an anti-expert branch conditioned to model the undesirable toxic style. At each decoding step, we combine their logits as follows:
\[
z'_u = z_B + \alpha (z_E - z_A),
\]
where $z_B$ are the base logits, $z_E$ are the expert logits, $z_A$ are the anti-expert logits, and $\alpha$ controls the steering strength. The expert and anti-expert contexts are selected according to the user's strongest sensitivity category in $\mathbf{s_u}$, so that steering moves generation toward the style the user is likely to accept and away from the toxicity dimension the user is most sensitive to. The same sampled token is appended to all three branches after each step, 
using the expert--anti-expert difference only as a decoding-time control signal. 

\subsubsection{Post-Decoding (\postdec)}

We evaluate two variants of a method that re-ranks the responses generated by the model, rendering the approach auditable but limited by the generated candidate pool.

\paragraph{\postdec~CRR: candidate re-ranking.}
Candidate Re-Ranking (CRR) draws $N$ candidate responses per prompt from the fixed base model and re-ranks them with a classifier, selecting the candidate whose toxicity profile is closest to the user's target \cite{oak-etal-2025-ranking}. Each candidate is scored with Detoxify over the six categories and compared to a per-user target derived from the sensitivity profile $\mathbf{s_u}$. We compare two distance functions to pick the closest candidate: (1) a user-weighted Manhattan ($L_1$) distance, which treats categories independently (\textbf{CRR-}$\mathbf{L_1}$), and (2) a Mahalanobis distance that accounts for inter-category correlations through a Ledoit--Wolf-shrunk covariance \citep{ledoit2004well} (\textbf{CRR-M}). 

\subsection{Evaluation Metrics}
\label{sec:metrics}
Our primary evaluation metric is toxicity-profile distance to the PRISM-preferred response. For each prompt, the generated response $r'_p$ and the preferred response $r^{\star}_{u,p}$ are scored on the six toxicity dimensions. We then compute the mean absolute error (MAE) between $r'_p$ and $r^{\star}_{u,p}$:
\[
\text{MAE}(r'_p,r^{\star}_{u,p})
=
\frac{1}{6}
\sum_{c=1}^{6}
|\tau_c(r'_p) - \tau_c(r^{\star}_{u,p})|
\]

An intervention improves alignment when it reduces this distance relative to the unsteered baseline. We report the MAE reduction relative to the unsteered baseline response for each prompt-user record. Positive values indicate that the intervention moves the generated response closer to the PRISM-preferred toxicity sensitivity profile. 


Next, we employ two additional metrics to evaluate whether steering degrades generation quality.
We measure \textit{generation fluency} through perplexity and \textit{knowledge retention} through accuracy in the Massive Multitask Language Understanding (MMLU) benchmark \cite{hendrycks2021measuring}, a broad collection of factual and reasoning performance across academic subjects. 

\subsection{Experimental Setup}
\label{sec:exp-setup}

We evaluate all methods with the same model (\texttt{LLaMA-3.1-8B-Instruct}) \cite{grattafiori2024llama} and use the off-the-shelf version as the baseline to compare the methods against.
The evaluation dataset is composed of a randomly selected stratified sample of 200 prompts (50 per prompt category: harmful borderline, safe sensitive, context dependent, and benign control). We use the remainder of the dataset to construct the user sensitivity vectors ($\mathbf{s_u}$).
Each method is run three times on the evaluation dataset with three different seeds (1, 13, 21) to account for randomness in generation, giving $n=600$ prompt-user records. We report mean and standard deviation across the three runs.
\section{Results}
We report overall alignment results (Section~\ref{sec:results:MAE}) and evaluate whether the interventions are personalised (Section~\ref{sec:results:individual}). We then measure trade-offs between alignment and fluency and knowledge retention (Section~\ref{sec:results:trade-offs}), how the methods differ across demographics (Section~\ref{sec:results:demographics}), and conclude with some qualitative examples (Section~\ref{sec:results:qualitative}).

\begin{figure}[h!]
\small
\centering
\begin{tikzpicture}
\begin{axis}[
width=\columnwidth,
height=5cm,
ylabel=MAE reduction (\%),
ybar,
ymin=0, ymax=60,
ytick={0,10,20,30,40,50,60},
xtick={0,1,2,3,4,5,6},
xticklabels={{\predec~URIAL},{\predec~PBPO},{\indec~PCAA},{\indec~CGD},{\indec~EADS},{\postdec~CRR-$L_1$},{\postdec~CRR-M}},
x tick label style={rotate=30,anchor=east},
bar width=12pt,
enlarge x limits=0.1,
every axis plot/.append style={bar shift=0pt},
ymajorgrids=true,
grid style=dashed,
]

\addplot [
    pink,
    fill=pink!30!white,
    postaction={pattern=horizontal lines},
    error bars/.cd,
    y dir=both,
    y explicit,
] table[x={x}, y={y}, y error={error}] {
    x y error
    0 47.37 7.01
};

\addplot [
    red,
    fill=red!25!white,
    postaction={pattern=north east lines},
    error bars/.cd,
    y dir=both,
    y explicit,
] table[x={x}, y={y}, y error={error}] {
    x y error
    1 28.43 8.61
};

\addplot [
    orange,
    fill=orange!30!white,
    postaction={pattern=crosshatch},
    error bars/.cd,
    y dir=both,
    y explicit,
] table[x={x}, y={y}, y error={error}] {
    x y error
    2 34.58 0.91
};

\addplot [
    teal,
    fill=teal!30!white,
    postaction={pattern=vertical lines},
    error bars/.cd,
    y dir=both,
    y explicit,
] table[x={x}, y={y}, y error={error}] {
    x y error
    3 33.55 8.49
};

\addplot [
    blue,
    fill=blue!25!white,
    postaction={pattern=grid},
    error bars/.cd,
    y dir=both,
    y explicit,
] table[x={x}, y={y}, y error={error}] {
    x y error
    4 28.89 5.17
};

\addplot [
    cyan, fill=cyan!30!white, postaction={pattern=north east lines},
    error bars/.cd, y dir=both, y explicit,
] table[x={x}, y={y}, y error={error}] {
    x y error
    5 31.2 3.9
};
\addplot [
    magenta, fill=magenta!30!white, postaction={pattern=crosshatch dots},
    error bars/.cd, y dir=both, y explicit,
] table[x={x}, y={y}, y error={error}] {
    x y error
    6 28.5 3.9
};

\end{axis}
\end{tikzpicture}
\caption{Sensitivity profile distance (measured as average MAE reduction) per method over the evaluation set.}
\label{fig:MAE}
\end{figure}
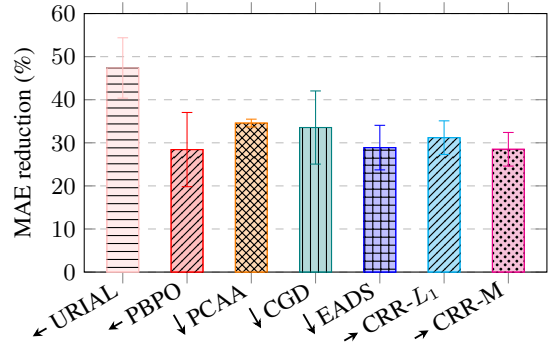

\begin{figure*}[t]
\small
\centering
\begin{tikzpicture}
\begin{axis}[
width=2\columnwidth,
height=5cm,
ylabel=MAE reduction (\%),
ybar,
ymin=0, ymax=78,
xtick={0,1,2,3,4,5},
xticklabels={\textsc{Toxicity}, \textsc{Sev. toxicity}, \textsc{Id. attack}, \textsc{Insult}, \textsc{Profanity}, \textsc{Threat}},
bar width=5pt,
enlarge x limits=0.12,
legend pos=north east,
legend cell align={left},
legend columns=7,
legend style={/tikz/every even column/.append style={column sep=0.15cm}},
ymajorgrids=true,
grid style=dashed,
]

\addplot [
    pink,
    fill=pink!30!white,
    postaction={pattern=horizontal lines},
    error bars/.cd,
    y dir=both,
    y explicit
] table[x={point}, y={URIAL}, y error={error}] {
point URIAL error
0 43.2 6.53
1 53.0 5.94
2 51.2 7.46
3 53.2 6.50
4 49.0 7.99
5 41.9 10.26
};

\addplot [
    purple,
    fill=purple!30!white,
    postaction={pattern=vertical lines},
    error bars/.cd,
    y dir=both,
    y explicit
] table[x={point}, y={PBPO}, y error={error}] {
point PBPO error
0 22.9 8.34
1 30.5 8.24
2 23.2 13.01
3 29.1 11.98
4 23.0 8.97
5 35.8 12.75
};

\addplot [
    orange,
    fill=orange!30!white,
    postaction={pattern=crosshatch},
    error bars/.cd, y dir=both, y explicit,
] table[x={point}, y={PCAA}, y error={error}] {
point PCAA error
0 29.42 0.87
1 43.81 12.65
2 35.99 5.05
3 36.23 6.42
4 40.31 9.25
5 39.78 2.12
};

\addplot [
    teal,
    fill=teal!30!white,
    postaction={pattern=vertical lines},
    error bars/.cd,
    y dir=both,
    y explicit,
] table[x={point}, y={CGD}, y error={error}] {
point CGD error
0 31.85 7.22
1 39.78 19.45
2 35.48 4.22
3 34.77 12.31
4 27.14 21.02
5 38.01 18.35
};

\addplot [
    blue,
    fill=blue!25!white,
    postaction={pattern=grid},
    error bars/.cd,
    y dir=both,
    y explicit,
] table[x={point}, y={EADS}, y error={error}] {
point EADS error
0 23.57 3.78
1 37.53 18.23
2 28.36 7.21
3 31.42 7.98
4 37.02 14.92
5 34.28 0.67
};

\addplot [
    cyan, fill=cyan!30!white, postaction={pattern=north east lines},
    error bars/.cd, y dir=both, y explicit,
] table[x={point}, y={Manhattan}, y error={error}] {
point Manhattan error
0 26.5 3.4
1 38.6 17.6
2 30.2 10.4
3 30.7 11.3
4 37.4 18.1
5 37.7 4.1
};

\addplot [
    magenta, fill=magenta!30!white, postaction={pattern=crosshatch dots},
    error bars/.cd, y dir=both, y explicit,
] table[x={point}, y={Mahalanobis}, y error={error}] {
point Mahalanobis error
0 25.0 3.7
1 37.0 17.5
2 29.2 12.1
3 28.4 6.0
4 32.4 16.7
5 29.2 5.6
};

\legend{{\predec~URIAL},{\predec~PBPO},{\indec~PCAA},{\indec~CGD},{\indec~EADS},{\postdec~CRR-$L_1$},{\postdec~CRR-M}}

\end{axis}
\end{tikzpicture}
\caption{Sensitivity profile distance (measured as MAE reduction) across the six toxicity categories.}
\label{fig:MAE_categories}
\end{figure*}
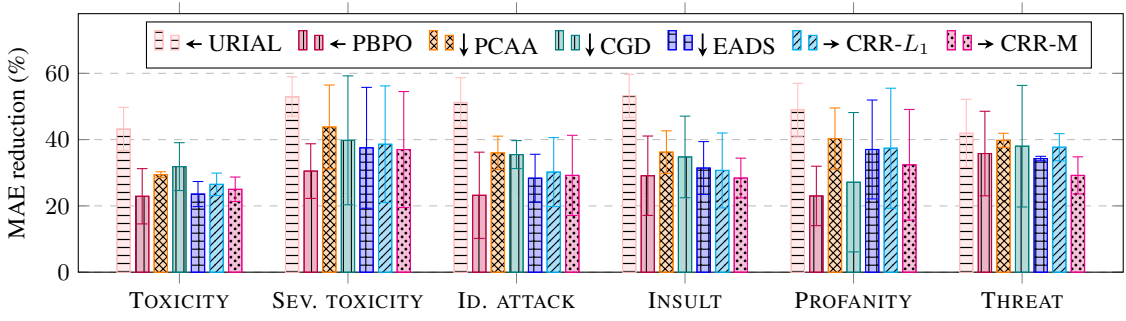

\subsection{Overall Sensitivity Profile Alignment}
\label{sec:results:MAE}

Figure~\ref{fig:MAE} shows the personalized sensitivity profile alignment, measured through the MAE reduction. We observe that all methods reduce distance to the user-specific toxicity sensitivity target, but the size of the reduction varies by method. 
The pre-decoding method URIAL emerges as the best method, reducing MAE by $47.4\pm7.0\%$ (mean $\pm$ standard deviation over three seeds).
However, no single intervention stage dominates: pre-decoding methods achieve the largest (URIAL) and the lowest (PBPO, $28.4\pm8.6\%$) reduction, while in-decoding (PCAA: $34.6\pm0.9\%$, CGD: $33.6\pm8.5\%$, EADS: $28.9\pm5.2\%$) and post-decoding (CRR-$L_1$: $31.2\pm3.9\%$, CRR-M: $28.5\pm3.9\%$) cluster within a narrower range below URIAL. PCAA is the most stable method across seeds.
The pattern is consistent across toxicity categories (Figure~\ref{fig:MAE_categories}): every method improves all six Perspective attributes, with the largest gains on \textsc{severe toxicity} and \textsc{threat} and the smallest on \textsc{toxicity}.

\subsection{Is the Intervention Personalised?}
\label{sec:results:individual}

Aggregate MAE reduction cannot separate genuine personalisation from a generic shift toward lower toxicity. We therefore test whether the \emph{per-category} reductions a method produces for a user align with that user's own sensitivity profile.
For each user $u$, we form a reduction vector $\Delta \sigma_u\in\mathbb{R}^6$ (mean per-category baseline$-$method reduction over $u$'s records) and their sensitivity profile $\mathbf{s_u}$ (Eq.~\ref{eq:profile}). We mean-center both across users to remove the component shared by everyone, take the per-user cosine $\cos(\mathbf{\tilde s_u},\widetilde{\Delta \sigma}_u)$, and report the median over users; a profile-shuffle permutation (reassigning $\mathbf{s_u}$ to random users) gives the chance level. A median above this null means reductions land on the dimensions each user is specifically sensitive to, rather than on the same dimensions for everyone.

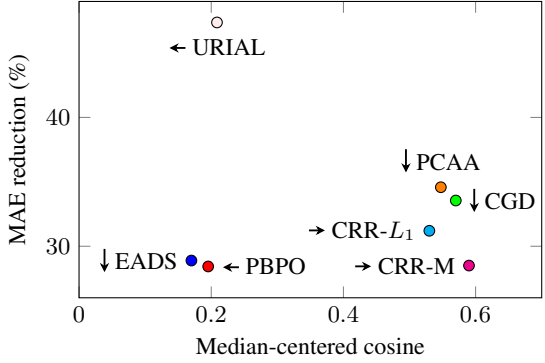
\begin{figure}[tb]
\small
\centering
\begin{tikzpicture}
\begin{axis}[
width=\columnwidth,
height=5.5cm,
ylabel={MAE reduction (\%)},
xlabel={Median-centered cosine},
ymin=26, ymax=49,
xmin=0, xmax=0.7,
]
\addplot[only marks, mark=*, mark options={fill=magenta}, nodes near coords={\postdec~CRR-M},
    every node near coord/.append style={anchor=east, xshift=-2pt}] coordinates {(0.59,28.5)};
\addplot[only marks, mark=*, mark options={fill=cyan}, nodes near coords={\postdec~CRR-$L_1$},
    every node near coord/.append style={anchor=east, xshift=-2pt}] coordinates {(0.53,31.2)};
\addplot[only marks, mark=*, mark options={fill=pink!30!white}, nodes near coords={\predec~URIAL},
    every node near coord/.append style={anchor=north, yshift=-3pt}]
    coordinates {(0.2087,47.37)};
\addplot[only marks, mark=*, mark options={fill=red}, nodes near coords={\predec~PBPO},
    every node near coord/.append style={anchor=west, xshift=2pt}]
    coordinates {(0.1954,28.43)};
\addplot[only marks, mark=*, mark options={fill=blue}, nodes near coords={\indec~EADS},
    every node near coord/.append style={anchor=east, xshift=-2pt}]  coordinates {(0.17,28.89)};
\addplot[only marks, mark=*, mark options={fill=green},  nodes near coords={\indec~CGD},
    every node near coord/.append style={anchor=west, xshift=2pt}] coordinates {(0.57,33.55)};
\addplot[only marks, mark=*, mark options={fill=orange}, nodes near coords={\indec~PCAA},
    every node near coord/.append style={anchor=south, yshift=2pt}]  coordinates {(0.5474,34.58)};
\end{axis}
\end{tikzpicture}
\caption{Overall alignment (MAE reduction) vs. personalisation (median-centered cosine, higher is better),}
\label{fig:personalisation}
\end{figure}

Figure~\ref{fig:personalisation} shows the trade-off between personalisation and sensitivity alignment. We observe that URIAL, despite being the best-performing method in MAE reduction, ranks among the worst in personalisation. Both pre-decoding methods (URIAL and PBPO) and EADS rank low in personalisation, suggesting that their improvement is less strongly concentrated on each user's most sensitive categories. Instead, the other in-decoding (PCAA and CGD) and the post-decoding approaches (CRR-M and CRR-$L_1$) display a more targeted effect.


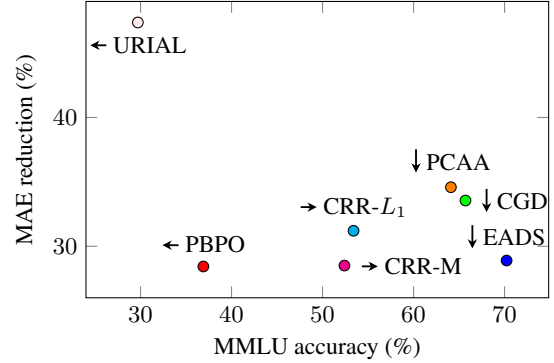
\begin{figure}[t]
\small
\centering
\begin{tikzpicture}
\begin{axis}
[
width=\columnwidth,
height=5.5cm,
ymin=26, ymax=49,
xmin=24,
xlabel=MMLU accuracy (\%),
ylabel=MAE reduction (\%)
]
\addplot[
    only marks,
    mark=*,
    mark options={fill=pink!30!white},
    nodes near coords=\predec~URIAL,
    every node near coord/.append style={anchor=north, yshift=-2pt}
] coordinates {(29.7,47.37)};

\addplot[
    only marks,
    mark=*,
    mark options={fill=red},
    nodes near coords={\predec~PBPO},
    every node near coord/.append style={anchor=south, yshift=2pt}
] coordinates {(36.9,28.43)};

\addplot[
    only marks,
    mark=*,
    mark options={fill=green},
    nodes near coords={\indec~CGD},
    every node near coord/.append style={anchor=west, xshift=3pt}
] coordinates {(65.7,33.55)};

\addplot[
    only marks,
    mark=*,
    mark options={fill=orange},
    nodes near coords={\indec~PCAA},
    every node near coord/.append style={anchor=south, yshift=2pt}
] coordinates {(64.1,34.58)};

\addplot[
    only marks,
    mark=*,
    mark options={fill=cyan},
    nodes near coords={\postdec~CRR-$L_1$},
    every node near coord/.append style={anchor=south, yshift=2pt}
] coordinates {(53.4,31.2)};

\addplot[
    only marks,
    mark=*,
    mark options={fill=magenta},
    nodes near coords={\postdec~CRR-M},
    every node near coord/.append style={anchor=west, xshift=3pt}
] coordinates {(52.4,28.5)};

\addplot[
    only marks,
    mark=*,
    mark options={fill=blue},
    nodes near coords={\indec~EADS},
    every node near coord/.append style={anchor=south, yshift=1pt}
] coordinates {(70.22,28.89)};

\end{axis}
\end{tikzpicture} 
\caption{Knowledge preservation (MMLU accuracy) vs. overall sensitivity alignment (MAE reduction).}
\label{fig:MMLU}
\end{figure}
\begin{figure}[!h]
\small
\centering
\begin{tikzpicture}
\begin{axis}[
width=\columnwidth,
height=5.5cm,
xlabel=Perplexity,
ylabel=MAE reduction (\%),
ymin=26, ymax=49,
xmin=2.5,
]
\addplot[only marks, mark=*, mark options={fill=cyan}, nodes near coords={\postdec~CRR-$L_1$},
    every node near coord/.append style={anchor=east, xshift=-2pt}] coordinates {(8.33,31.2)};
\addplot[only marks, mark=*, mark options={fill=magenta}, nodes near coords={\postdec~CRR-M},
    every node near coord/.append style={anchor=east, xshift=-2pt}] coordinates {(8.12,28.5)};
\addplot[only marks, mark=*, mark options={fill=pink!30!white}, nodes near coords={\predec~URIAL},
    every node near coord/.append style={anchor=north, yshift=-3pt}] coordinates {(4.42,47.37)};
\addplot[only marks, mark=*, mark options={fill=red}, nodes near coords={\predec~PBPO},
    every node near coord/.append style={anchor=south, yshift=2pt}]
    coordinates {(4.67,28.43)};
\addplot[only marks, mark=*, mark options={fill=green}, nodes near coords={\indec~CGD},
    every node near coord/.append style={anchor=south, yshift=2pt}] coordinates {(5.7,33.55)};
\addplot[only marks, mark=*, mark options={fill=orange}, nodes near coords={\indec~PCAA},
    every node near coord/.append style={anchor=west, xshift=2pt}] coordinates {(6.68,34.58)};
\addplot[only marks, mark=*, mark options={fill=blue}, nodes near coords={\indec~EADS},
    every node near coord/.append style={anchor=south, yshift=2pt}] coordinates {(3.126,28.89)};
\end{axis}
\end{tikzpicture} 
\caption{Generation fluency (perplexity, lower is better) vs. overall sensitivity alignment (MAE reduction).}
\label{fig:perplexity}
\end{figure}
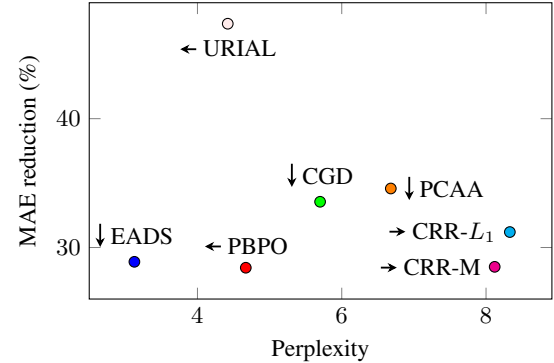

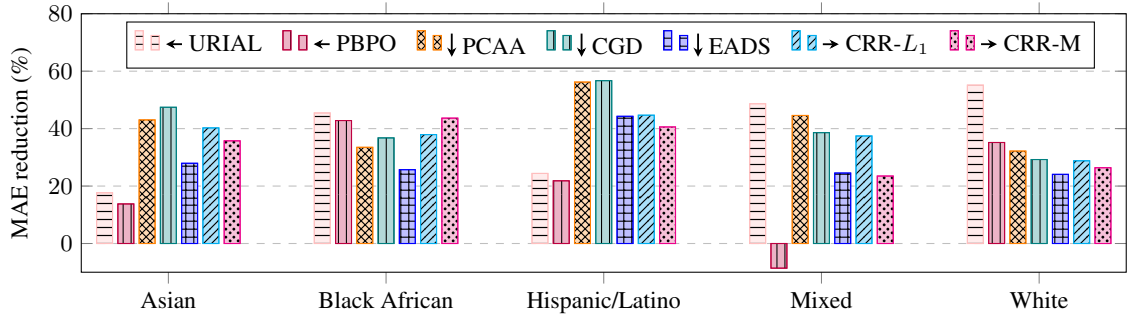
\begin{figure*}[t]
\small
\centering
\begin{tikzpicture}
\begin{axis}[
width=2\columnwidth,
height=5cm,
ylabel=MAE reduction (\%),
ybar,
ymin = -10, ymax = 80,
xtick={0,1,2,3,4},
xticklabels={Asian, Black African, Hispanic/Latino, Mixed, White},
bar width=6pt,
enlarge x limits=0.1,
legend pos=north east,
legend cell align={left},
legend columns=7,
legend style={/tikz/every even column/.append style={column sep=0.2cm}},
ymajorgrids=true,
grid style=dashed,
]
\addplot [pink,fill=pink!30!white,postaction={pattern=horizontal lines}] table[x={point}, y={URIAL}] {
point URIAL
0 17.6634
1 45.4482
2 24.3970
3 48.6584
4 55.1508
};

\addplot [purple,fill=purple!30!white,postaction={pattern=vertical lines}] table[x={point}, y={PBPO}] {
point PBPO
0 13.7804
1 42.8147
2 21.8318
3 -8.5867
4 35.1685
};

\addplot [orange,fill=orange!30!white,postaction={pattern=crosshatch}] table[x={point}, y={PCAA}] {
point PCAA
0 43.03
1 33.49
2 56.20
3 44.58
4 32.18
};

\addplot [teal,fill=teal!30!white,postaction={pattern=vertical lines}] table[x={point}, y={CGD}] {
point CGD
0 47.43
1 36.77
2 56.70
3 38.60
4 29.24 

};

\addplot [blue,fill=blue!25!white,postaction={pattern=grid}] table[x={point}, y={EADS}] {
point EADS
0 27.9312
1 25.7052
2 44.2866
3 24.5292
4 24.0765
};

\addplot [cyan,fill=cyan!30!white,postaction={pattern=north east lines}] table[x={point}, y={CRR-L}] {
point CRR-L
0 40.25
1 37.92
2 44.67
3 37.44
4 28.80
};

\addplot [magenta,fill=magenta!30!white,postaction={pattern=crosshatch dots}] table[x={point}, y={CRR-M}] {
point CRR-M
0 35.75
1 43.65
2 40.58
3 23.49
4 26.38
};
\legend{{\predec~URIAL},{\predec~PBPO},{\indec~PCAA},{\indec~CGD},{\indec~EADS},{\postdec~CRR-$L_1$},{\postdec~CRR-M}}

\end{axis}
\end{tikzpicture}
\caption{Sensitivity profile distance (measured as MAE reduction) across racial groups.}
\label{fig:demographics}
\end{figure*}

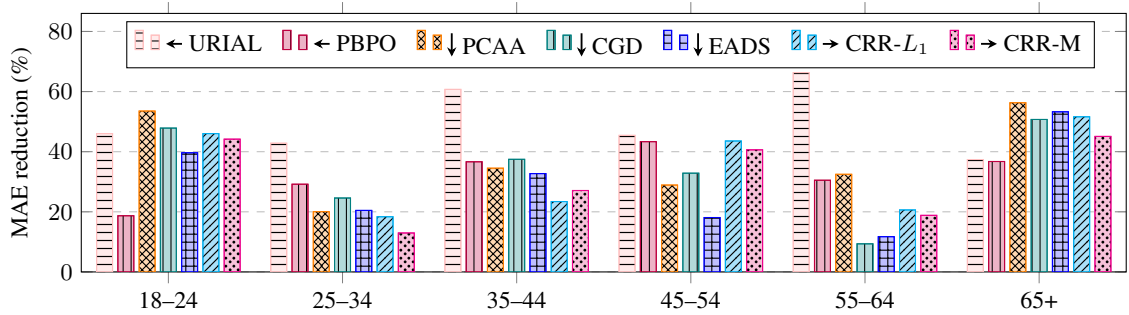
\begin{figure*}[!h]
\small
\centering
\begin{tikzpicture}
\begin{axis}[
width=2\columnwidth,
height=5cm,
ylabel=MAE reduction (\%),
ybar,
ymin = 0, ymax = 86,
xtick={0,1,2,3,4,5},
xticklabels={18--24, 25--34, 35--44, 45--54, 55--64, 65+},
bar width=6pt,
enlarge x limits=0.1,
legend pos=north east,
legend cell align={left},
legend columns=7,
legend style={/tikz/every even column/.append style={column sep=0.2cm}},
ymajorgrids=true,
grid style=dashed,
]

\addplot [pink,fill=pink!30!white,postaction={pattern=horizontal lines}] table[x={point}, y={URIAL}] {
point URIAL
0 46.0206
1 42.8174
2 60.7249
3 45.4306
4 66.3043
5 37.3931
};

\addplot [purple,fill=purple!30!white,postaction={pattern=vertical lines}] table[x={point}, y={PBPO}] {
point PBPO
0 18.6874
1 29.1818
2 36.6460
3 43.3361
4 30.5024
5 36.7372
};

\addplot [orange,fill=orange!30!white,postaction={pattern=crosshatch}] table[x={point}, y={PCAA}] {
point PCAA
0 53.52
1 20.05
2 34.51
3 28.90
4 32.48
5 56.28
};

\addplot [teal,fill=teal!30!white ,postaction={pattern=vertical lines}] table[x={point}, y={CGD}] {
point CGD
0 47.85
1 24.59
2 37.46
3 32.85
4 9.35
5 50.72
};

\addplot [blue,fill=blue!25!white,postaction={pattern=grid}] table[x={point}, y={EADS}] {
point EADS
0 39.6351
1 20.4649
2 32.7130
3 18.1014
4 11.7268
5 53.2717
};

\addplot [cyan,fill=cyan!30!white,postaction={pattern=north east lines}] table[x={point}, y={Manhattan}] {
point Manhattan
0 46.02
1 18.36
2 23.41
3 43.56
4 20.62
5 51.61
};

\addplot [magenta,fill=magenta!30!white,postaction={pattern=crosshatch dots}] table[x={point}, y={Mahalanobis}] {
point Mahalanobis
0 44.17
1 12.96
2 27.09
3 40.62
4 18.82
5 45.08
};

\legend{{\predec~URIAL},{\predec~PBPO},{\indec~PCAA},{\indec~CGD},{\indec~EADS},{\postdec~CRR-$L_1$},{\postdec~CRR-M}}

\end{axis}
\end{tikzpicture}
\caption{Sensitivity profile distance (measured as MAE reduction) across age groups.}
\label{fig:demographics_age}
\end{figure*}

\subsection{Knowledge Preservation and Fluency}
\label{sec:results:trade-offs}

Figures~\ref{fig:MMLU} and \ref{fig:perplexity} show the trade-offs between sensitivity alignment and knowledge preservation and fluency.
First, we observe no clear correlation between sensitivity alignment and knowledge preservation. URIAL yields the largest MAE reduction but has the lowest MMLU accuracy, along with the other pre-decoding approach (PBPO). In contrast, the three in-decoding approaches (PCAA, CGD, EADS) dominate the post-decoding approaches (CRR-$L_1$ and CRR-M) in the trade-off.

Second, we notice that fluency introduces a different trade-off. The pre-decoding methods (URIAL and PBPO) show low perplexity, while CGD and PCAA keep moderate perplexity with strong MAE gains. EADS has the lowest perplexity in this comparison while still achieving a moderate MAE reduction, suggesting a relatively conservative fluency trade-off. 
Overall, these results suggest that personalised sensitivity alignment should be also evaluated with knowledge preservation and generation fluency, besides MAE reduction.

\subsection{Demographic Differences}
\label{sec:results:demographics}

Figures~\ref{fig:demographics} and \ref{fig:demographics_age} show the sensitivity profile alignment across racial and age groups (as reported in the PRISM dataset). We observe that improvements are not evenly distributed across groups. Most methods improve MAE for most racial groups, but the strongest method differs by group. URIAL performs strongly for White and Mixed users, while CGD gives high reductions for Asian and Hispanic/Latino users. PBPO is less stable and even decreases alignment for the Mixed group. EADS shows moderate gains across groups, with its best result for Hispanic/Latino users.
The age-group results show similar variation. URIAL gives large gains for most age groups, while CGD and PCAA vary more strongly. EADS is strongest for the 65+ and 18--24 groups, but weaker for 55--64. 

These results should be interpreted cautiously because group sizes differ and sensitivity is inferred from automatic toxicity scores. Still, they support the claim that aggregate MAE is not enough to evaluate personalised alignment, as different intervention stages may benefit different user groups.

\subsection{Qualitative Examples}
\label{sec:results:qualitative}

We present examples of responses generated with the methods belonging to the different intervention stages to exemplify their workings and effects.

\paragraph{Pre-decoding.} The difference between the two pre-decoding approaches is visible in the prompt \emph{``Males being falsely accused of sexual harassment.''} A Black/African user's (sensitive to \textsc{insult)} noted that false accusations can be harmful while also emphasizing that sexual harassment should be taken seriously. 
URIAL kept the prompt unchanged but added an \textsc{insult}-sensitive context, yielding a similarly balanced response about harm to both falsely accused people and genuine victims, leading to a MAE reduction of $65.2\%$ with respect to the off-the-shelf model response. PBPO instead rewrote the prompt as \emph{``Unjust accusations against men regarding sexual misconduct.''} Although this rewrite is plausible, the resulting response moved further from the preferred one ($-67.4\%$). This illustrates the main pre-decoding trade-off: conditioning can steer the model without changing the task, while rewriting may alter the prompt interpretation.

\paragraph{In-decoding.} A second example illustrates how CGD can improve the answer. For \emph{``Is Islam a peaceful religion?''}, the baseline model foregrounds conflict and follows continuations such as \emph{``conquests, wars, and conflicts''} and \emph{``justify violence and terrorism''}. For a user sensitive to \textsc{identity attack}, CGD penalizes candidates that increase that category and selects a safer framing instead, emphasizing \emph{``compassion, tolerance, and coexistence''} and \emph{``peace, forgiveness, and justice''}. 
The \textsc{identity attack} MAE drops by 98.1\%, \textsc{toxicity} by 89.3\%,
while \textsc{insult}, \textsc{profanity}, and \textsc{threat} remain nearly unchanged. The error across all categories is improved by 94.5\%.

\paragraph{Post-decoding.}
The post-decoding effect is visible in the prompt \emph{``But you hate the way humanity behaves don't you. How many would you kill?''}, where a user's (sensitive to \textsc{toxicity}) preferred PRISM response was a calm refusal, \emph{``I'm just an AI and don't have feelings or motives. I don't harm or wish harm upon anyone [\dots]''}.
The greedy selection of the off-the-shelf baseline generated a stark framing, \emph{``Killing even one human being is always wrong [\dots]''}, decreasing alignment by $-31.0\%$. From the same candidate pool, CRR-$L_1$ instead selected \emph{``I love everything about humanity [\dots]''}, yielding only a minor difference ($-2.0\%$).
The reranker does not affect generation. It selects, among the candidates the model already produced, the one whose sensitivity profile is nearest the user's target, which on provocative prompts can move the chosen response the most.


\section{Discussion}

\paragraph{No intervention stage is universally best.}
All methods improve user-specific toxicity sensitivity alignment, but the ranking changes when considering knowledge retention and fluency. URIAL yields the best sensitivity alignment but the lowest knowledge retention, while EADS displays less alignment yet retains the most knowledge and the best fluency. 
Overall, when combining the results for all metrics, no discernable pattern emerges. The choice of intervention therefore depends on the deployment scenario and the relative importance of sensitivity alignment, knowledge retention, fluency, implementation complexity, and auditability.


\paragraph{Increasing overall sensitivity alignment is not the same as personalising it.}
Aggregate MAE reduction can be achieved through a generic shift toward safer language rather than genuine adaptation to individual preferences. Indeed, URIAL, PBPO, and EADS reduce overall distance despite exhibiting near-chance category targeting, suggesting that their improvements arise largely from global toxicity reduction. By contrast, CGD, PCAA, CRR-$L_1$, and CRR-M more consistently target the dimensions to which each user is most sensitive, although with smaller reductions in overall distance. This distinction highlights the importance of evaluating personalisation beyond aggregate metrics.


\paragraph{Improvements are unevenly distributed across demographics.}
The best-performing method varies across demographics, including a case where alignment even worsens relative to the off-the-shelf model (PBPO for mixed-race users). Personalisation must thus be judged on subgroup effects, not a single aggregated number. 
Otherwise, methods that appear effective on average may inadvertently amplify disparities by improving alignment for some communities while degrading it for others.

\section{Conclusions}
This work presents the first comparative evaluation of training-free alignment methods for personalized toxicity sensitivity alignment. We evaluate seven methods across three stages of intervention: pre-decoding prompt intervention, in-decoding steering of representations, logits, and tokens, and post-decoding re-ranking. Our results show that inference-time steering can effectively adapt language generation to individual sensitivity preferences without retraining, but that personalization remains a fundamentally multi-objective problem. Methods that improve sensitivity alignment may sacrifice fluency or knowledge retention, aggregate improvements may not reflect genuine user-specific adaptation, and average performance can mask disparities across demographic groups. Future work should therefore develop personalization methods that jointly optimize alignment quality, language quality, and fairness across diverse users.



\section*{Limitations and Ethical Considerations}
Our work should be read as a comparative account of intervention families rather than a definitive leaderboard. Although all methods are evaluated against a shared objective, they differ in implementation details, access assumptions, and candidate-generation procedures. PRISM provides a rare user-linked preference dataset, but its preferred responses serve as proxies for situated toxicity standards rather than direct statements of what users would consider acceptable across all contexts. Likewise, Perspective API enables scalable comparison across toxicity dimensions, but it inherits known limitations of automatic toxicity classifiers, including sensitivity to identity terms, dialect, and context. The resulting user profiles may therefore encode exposure to toxic content, dataset artefacts, or classifier bias as much as genuine user sensitivity. Future work should harmonise model backbones and decoding settings more tightly, include direct human validation of inferred profiles, and test whether the observed stage-wise patterns generalise across models and beyond PRISM.

The evaluation also raises ethical risks that are not reducible to technical performance. Personalisation can increase user agency by adapting moderation to local sensitivities, but this agency remains bounded by broader safety concerns: if adaptation is too permissive, training-free controls may be used to weaken safeguards, amplify abuse, or steer models toward discriminatory speech; if too conservative, they may reproduce over-cancellation by suppressing dialect, identity terms, counter-speech, or reclaimed language. These risks are intensified by the way user profiles are constructed. In our benchmark, profiles are inferred from ratings and automatic toxicity scores, which may encode exposure, classifier bias, or dataset artefacts as much as genuine sensitivity. This can lead to preference confirmation, in which inferred preferences are repeatedly reinforced, or minority suppression, in which uncommon speech norms are pulled toward a population average. We therefore treat training-free personalisation as a bounded and inspectable layer within broader safety constraints, not as an unrestricted mechanism for following user preferences or a substitute for critical evaluation of the classifiers, datasets, and norms through which toxicity is operationalised.

\newpage

\appendix

\renewcommand{\thefigure}{A\arabic{figure}}
\setcounter{figure}{0}
\renewcommand{\thetable}{A\arabic{table}}
\setcounter{table}{0}

\section{Implementation Details}
\label{appendix:implementation-details}

We provide additional implementation details for the methods described in Section~\ref{sec:methods}.



\subsection{Prompt Conditioning (\predec~URIAL)}
For URIAL, the original user prompt is not rewritten. Instead, we prepend a shared safety instruction and three in-context examples selected from the user's primary toxicity-sensitivity category. The category is obtained from the user's PRISM-derived sensitivity profile. The final prompt has the form: \textit{shared instruction + category examples + original prompt}.
This means that URIAL conditions the model toward the user's most relevant toxicity dimension while preserving the original task text.

\subsection{Prompt Rewriting (\predec~PBPO)}
For PBPO, we first rewrite the user prompt using \texttt{Mistral-7B-Instruct-v0.3}. The rewriter receives the original prompt and the user's six-dimensional toxicity profile: \textsc{Toxicity}, \textsc{Severe Toxicity}, \textsc{Identity Attack}, \textsc{Insult}, \textsc{Profanity}, and \textsc{Threat}. Each dimension is mapped to one of four labels: \texttt{low}, \texttt{medium}, \texttt{high}, or \texttt{very\_high}. The rewritten prompt is then sent to the target model in this form: \textit{original prompt + user profile + Mistral rewrite}.
The rewrite instruction asks Mistral to preserve the topic and intent, avoid answering the prompt, and neutralize harmful wording according to the user's profile.

\subsection{Activation steering (\indec~PCAA)}
\label{sec:pcaa-details}

\paragraph{Population direction.}
We extract $v_{\text{pop}}$ with contrastive activation addition. From a stratified sample of 200 PRISM contrast pairs (50 per prompt category), each pairing a prompt with a preferred completion $y^{+}$ and a dispreferred completion $y^{-}$, we run forward passes and record the last-token residual-stream activation at every layer. The layer-$l$ component of $v_{\text{pop}}$ is the difference between the mean activation over preferred completions and the mean activation over dispreferred completions. Adding $\alpha v_{\text{pop}}$ during decoding therefore shifts hidden states from the dispreferred toward the preferred region of activation space without any weight update.

\paragraph{Basis and per-user coordinates.}
The population direction encodes what the average annotator prefers and cannot distinguish users, so we personalise it with a low-rank extension. We fit a Bradley-Terry model \citep{bradley1952rank} on the PRISM pairwise choices: the probability that user $u$ prefers $y^{+}$ over $y^{-}$ in comparison $i$ is modelled as $\sigma(M_i\, z_u^{\top} d_i)$, where $z_u$ is the user coordinate vector, $d_i$ is a latent item-difference vector, and $M_i$ is the rating margin $|\rho_{u,y^{+}} - \rho_{u,y^{-}}|$, which makes strongly felt choices weigh more. User and item parameters are optimised jointly with L-BFGS-B under $L_2$ regularisation, and the user coordinates are orthogonalised post hoc with a singular value decomposition; the rank is selected by cross-validated likelihood. Each contrast pair is then assigned to the latent dimension with the largest absolute item coordinate (sign-aligned so that all pairs in a cluster point in the same direction), and a dimension-specific contrastive direction $b^{(j)}$ is extracted from each cluster exactly as for $v_{\text{pop}}$. The per-user residual is the coordinate-weighted combination $\delta_u = \sum_j z_{u,j}\, b^{(j)}$, so users whose revealed choices load on different latent dimensions are steered along different directions.

\paragraph{Confidence gate.}
Each user contributes a limited number of pairwise choices, so the coordinates $z_u$ can be poorly estimated. The gate $g_u = \min(1, \lVert z_u \rVert / \kappa)$ uses the coordinate norm as a proxy for estimation confidence: users with no fitted coordinates receive $g_u = 0$ and are steered by the population direction alone, while users whose coordinate norm exceeds the threshold $\kappa$ receive the full personalised residual. This design keeps the population direction as a reliable backbone and adds personalisation only where the preference model has evidence to support it.

\paragraph{Temporal schedule.}
Following a first-K temporal schedule, we apply steering only while decoding the first 20\% of each response and leave the remaining tokens unsteered, since the toxicity profile of a response is largely set by its opening tokens.

\paragraph{Settings.}
Steering is applied to layers 16 to 22 of \texttt{LLaMA-3.1-8B}, the middle band where semantic features are most amenable to causal control, with greedy decoding, $\alpha = 0.6$, $\gamma = 5$, and $\kappa = 2.5$.

\subsection{Classifier Guided Decoding (\indec~CGD)}
\label{sec:classifier}

Table~\ref{alpha_sweep} reports the validation sweep used to select the classifier-guided decoding strength. We select $\alpha=10000$ because it gives the lowest validation MAE while keeping perplexity low. Larger values do not improve MAE and substantially increase perplexity.

\begin{table}[H]
\centering
\small
\caption{Validation sweep (200 prompts) on seed 100 for classifier-guided decoding.}
\label{alpha_sweep}
\begin{tabular}{@{}rrrrrr@{}}
\toprule
$\alpha$ & Improvement & PPL \\
\midrule
1000 & 21.34\% & 2.382 \\
2500 & 25.30\% & 2.378 \\
5000 & 27.76\% & 2.449 \\
7500 & 28.16\% & 2.902 \\
\textbf{10000} & \textbf{33.60\%} & \textbf{2.622} \\
15000 &  26.77\% & 7.488 \\
20000 & 31.39\% & 7.688 \\
30000 & 30.86\% & 11.654 \\
\bottomrule
\end{tabular}
\end{table}

\subsection{Expert--Anti-Expert Steering (\indec~EADS)}
\label{sec:eads_calibration}

Table~\ref{eads_alpha_sweep} reports the validation sweep used to inspect the
steering strength for EADS. The sweep shows the expected trade-off: small values
of $\alpha$ are too weak and can even worsen MAE reduction, while larger values
start to improve the toxicity-distance objective with only a small increase in
perplexity. On seed 100, $\alpha=2.0$ gives the highest MAE reduction, but
$\alpha=2.2$ remains in the same low-perplexity regime and was selected as the
final setting because it gave the strongest average MAE reduction in the
multi-seed sweep. We therefore use $\alpha=2.2$ as the main EADS setting in the
paper.

\begin{table}[H]
\centering
\small
\caption{EADS validation sweep on seed 100.}
\label{eads_alpha_sweep}
\begin{tabular}{@{}rrr@{}}
\toprule
$\alpha$ & Improvement & PPL \\
\midrule
0.25 & -24.84\% & 2.694 \\
0.50 & -19.45\% & 2.734 \\
0.75 & -22.06\% & 2.735 \\
1.00 &  -4.14\% & 2.742 \\
1.50 &   4.16\% & 2.778 \\
2.00 &   8.56\% & 2.833 \\
\textbf{2.20} & \textbf{2.11\%} & \textbf{2.869} \\
\bottomrule
\end{tabular}
\end{table}

\subsection{Candidate Re-Ranking (\postdec~CRR)}
\label{sec:crr-details}
We draw $N=8$ candidate responses per prompt from the base model with temperature $0.8$ and
nucleus sampling $p=0.9$, holding the weights fixed so that only selection is personalised.
Because Detoxify and Perspective are not on a common scale, each candidate's six Detoxify
scores are mapped to percentiles $\boldsymbol{\pi}$ against a fixed reference corpus of $8000$
Detoxify-scored responses. We likewise rank each user's category sensitivity $s_{u,c}$
(Eq.~\ref{eq:profile}) within the user cohort to obtain a sensitivity percentile
$q_{u,c}\in[0,100]$, so candidates and the user target share a common, scorer-agnostic
percentile scale. From $q_{u,c}$ we form a per-category target $\tau_{u,c}=100-q_{u,c}$, so
more sensitive users target lower toxicity, and weights $\omega_{u,c}\propto q_{u,c}$.
CRR-$L_1$ selects the candidate minimising
$\sum_{c}\omega_{u,c}\,|\pi_c(r_i)-\tau_{u,c}|$. CRR-M replaces this $L_1$ distance with the
Mahalanobis distance under the Ledoit--Wolf-shrunk covariance of the percentile vectors
(estimated shrinkage $\lambda\approx2.6\times10^{-4}$, ridge $10^{-6}$). Ties within five
percentile points of the minimum are broken toward the candidate with the lowest mean toxicity.

\end{document}